\documentclass[sigconf]{acmart}

\setcopyright{none}                    
\renewcommand\footnotetextcopyrightpermission[1]{} 
\acmConference{}{}{}                   
\acmBooktitle{}                        
\acmDOI{} \acmISBN{} \acmPrice{}       
\usepackage{fancyhdr}
\makeatletter
\fancypagestyle{plain}{
  \fancyhf{}%
}
\makeatother

\AtBeginDocument{%
  }

\usepackage{multirow}
\usepackage{colortbl}
\definecolor{green}{RGB}{210, 235, 210}
\definecolor{pink}{RGB}{250, 225, 225}
\definecolor{Beige}{RGB}{240, 235, 210}
\definecolor{ImproveColor}{RGB}{200, 200, 200}
\usepackage{amsmath,amsfonts}
\usepackage{amsthm}
\usepackage{booktabs}
\usepackage[linesnumbered, ruled, vlined]{algorithm2e}
\usepackage{enumitem}

\begin{document}

\title{Forecasting at Full Spectrum: Holistic Multi-Granular Traffic Modeling under High-Throughput Inference Regimes}

\author{Zhaoyan Wang}
\authornote{Corresponding author.}
\affiliation{%
  \institution{School of Computing, Korea Advanced Institute of Science and Technology (KAIST)}
  \country{}
}
\email{zhaoyan123@kaist.ac.kr}

\author{Xiangchi Song}
\affiliation{%
  \institution{School of Computing, Korea Advanced Institute of Science and Technology (KAIST)}
  \country{}
}
\email{xcsong@kaist.ac.kr}

\author{In-Young Ko}
\affiliation{%
  \institution{School of Computing, Korea Advanced Institute of Science and Technology (KAIST)}
  \country{}
}
\email{iko@kaist.ac.kr}


\begin{abstract}
  Notably, current intelligent transportation systems rely heavily on accurate traffic forecasting and swift inference provision to make timely decisions. While Graph Convolutional Networks (GCNs) have shown benefits in modeling complex traffic dependencies, the existing GCN-based approaches cannot fully extract and fuse multi-granular spatiotemporal features across various spatial and temporal scales sufficiently in a complete manner, proven to yield less accurate results. Besides, as extracting multi-granular features across scales has been a promising strategy across domains such as computer vision, natural language processing, and time-series forecasting, pioneering studies have attempted to leverage a similar mechanism for spatiotemporal traffic data mining. However, additional feature extraction branches introduced in prior studies critically increased model complexity and extended inference time, making it challenging to provide fast forecasts. In this paper, we propose MultiGran-STGCNFog, an efficient fog distributed inference system with a novel traffic forecasting model that employs multi-granular spatiotemporal feature fusion on generated dynamic traffic graphs to fully capture interdependent traffic dynamics. The proposed scheduling algorithm GA-DPHDS, optimizing layer execution order and layer-device scheduling scheme simultaneously, contributes to considerable inference throughput improvement by coordinating heterogeneous fog devices in a pipelined manner. Extensive experiments on real-world datasets demonstrate the superiority of the proposed method over selected GCN baselines.
\end{abstract}





\begingroup
\setlength{\bigskipamount}{2em} 
\maketitle
\endgroup

\pagestyle{plain}     
\thispagestyle{plain} 

\section{Introduction}
Nowadays, as a core pillar of smart city paradigm, intelligent transportation systems play a pivotal role in enhancing urban mobility, reducing congestion, and improving transportation efficiency. The large volume of traffic data has shifted attention toward data analysis algorithms and corresponding servings to enhance transportation system intelligence~\cite{zantalis2019review}. Traditional centralized transportation systems are limited by high latency and weak scalability. Against this backdrop, fog computing, which processes data on nearer nodes close to the data source, has emerged as a more efficient and scalable solution. Collected traffic data needs to be analyzed efficiently to predict future traffic conditions as accurate predictions benefit services like transportation optimization, navigation planning, and congestion prevention~\cite{li2022automated}. Consequently, traffic forecasting has become an essential task, focusing on predicting future road network patterns using historical observations~\cite{liu2023spatio}.

Embodied by advances across domains, the exploitation of multi-granular features at different scales has emerged as a promising mechanism for enhancing model expressiveness and robustness across a wide range of learning tasks, which demonstrates strong capability in capturing hierarchical structures and contextual dependencies from complex data. In computer vision, multi-scale designs are often realized through pyramid structures or parallel convolutional kernels, enabling models to detect objects and interpret scenes across spatial resolutions~\cite{Lin_2017_CVPR,chen2018encoder,zhao2017pyramid,szegedy2017inception}. In natural language processing, hierarchical representations can be constructed by integrating local token-level dependencies with broader document-level semantics, via attention mechanisms that operate across multiple textual spans~\cite{dai2019transformer,beltagy2020longformer,zhang2019ernie,jia2019document,jin2020multi}. In the realm of time-series analysis, models benefit from capturing both short-term fluctuations and long-term temporal trends, typically through multi-resolution temporal blocks and frequency-domain decomposition~\cite{lim2021temporal,zhou2021informer,wu2022timesnet,cui2016multi}. In line with other research domains, the exploration of multi-scale mechanisms has commenced in traffic data mining.

Existing traffic forecasting models primarily design complex Spatial-Temporal (S-T) blocks but lack a structured approach to extract and integrate hierarchical, multi-granular spatiotemporal features across spatial and temporal dimensions. Although a few works explored multi-granular spatial~\cite{wang2021mt,guo2021hierarchical,chen2024urban,li2024seeing} or temporal features~\cite{dan2024bygcn,wang2022spatiotemporal,huo2023hierarchical,guo2019attention} independently, traffic forecasting involves both spatial and temporal characteristics mining. Seamlessly integrating both is crucial for accurate forecasts and remains unexplored.

Besides, previous studies manipulating multi-scale traffic data introduced multiple feature extraction branches to capture features of different granularities. This significantly increased model complexity and led to extended inference time, making it challenging for Graph Neural Network (GNN) serving where efficiency and low latency are critical. No solutions were proposed in these earlier works to address this issue and no framework was specifically designed for accelerating multi-granular spatiotemporal traffic forecasting.

While GCN-based models dominate spatiotemporal forecasting, transformer-based ones have recently shown efficacy in capturing long-term dependencies~\cite{jiang2023pdformer,liu2023spatio}. However, GCNs still remain irreplaceably unmatched in handling graph-structured spatial dependencies, and with carefully designed architectures, they are capable of achieving better performance than transformers~\cite{dan2024bygcn,zhang2024irregular}. In light of the above, we seek to answer two main research questions: 

\begin{enumerate}[label=\textbf{RQ\arabic*.}]
    \item Whether constructing a comprehensive multi-scale feature extraction and fusion mechanism—jointly designed across both spatial and temporal dimensions—is able to enhance the accuracy of GCN-based traffic forecasting?
    \item Can an inference-optimized framework benefit the inference throughput of multi-granular spatiotemporal traffic forecasting models, and in what manner does it achieve the improvement?
\end{enumerate}



Furthermore, designing accurate traffic forecasting models with swift inference still encounters several key challenges:

1) Traffic data exhibits highly dynamic and sophisticated patterns, making it difficult to model spatiotemporal correlations.

2) In GNN serving systems, efficiently leveraging available resources is essential for speeding up where communication latency becomes a major hurdle~\cite{zeng2022fograph}.

3) Optimizing distributed parallel inference schemes is increasingly complicated due to the exponentially-growing search space for the optimal layer execution order and layer allocation scheduling, together with layer input-output dependencies.

4) Mainstream distributed parallel frameworks assume even partitioning of stages, ignoring hardware heterogeneity. Real-word device clusters are often heterogeneous, and model layers cannot be fully in even distribution to ensure equal processing times.

To fill these gaps, we propose MultiGran-STGCNFog, an efficient GNN inference system that extracts and fuses multi-granular spatiotemporal features across various spatial and temporal scales. The distributed pipeline-parallel architecture across fog nodes enables MultiGran-STGCNFog to tackle the extended inference latency, supporting accurate and swift traffic forecasting. 

Our contributions can be summarized as follows:
\sloppy\begin{itemize}
\item A novel GCN, \underline{\textbf{Multi-Gran}}ularity \underline{\textbf{S}}patio\underline{\textbf{t}}emporal \underline{\textbf{G}}raph \underline{\textbf{C}}onvolution \underline{\textbf{N}}etwork (\textbf{MultiGran-STGCN}), is proposed. It employs dynamic graph techniques and a multi-granular spatiotemporal feature fusion mechanism through Laplacian-driven hierarchical graph clustering and multi-scale temporal modeling to exploit multi-granular features.
\item To mitigate the prolonged inference time induced by additional feature extraction branches, we construct \textbf{MultiGran-STGCNFog} for inference acceleration, leveraging the resources of heterogeneous fog devices. It enables faster inference by pipeline parallelism, which allows immediate data transmission among fog nodes in proximity to avoid remote transfers.
\item We design a robust heterogeneous cross-device scheduling algorithm, \textbf{GA-DPHDS}, to optimize the layer execution order and layer-device scheduling scheme jointly, overcoming the search space explosion challenge. Experiment results show the necessity of layer execution order optimization, which is overlooked in previous pipeline parallelism works.
\item The proposed framework is tested with physical computing nodes and evaluated using three real-world traffic datasets. Up to \textbf{9.86\%} forecasting metric improvement and \textbf{2.43x} throughput improvement demonstrate its superior performance compared with selected baselines.
\end{itemize}

\section{Related Work} \label{sec:rework}
\subsection{GCNs: Spatiotemporal Traffic Forecasting}
When combining temporal modeling techniques like RNNs, GCN-based models are particularly effective at capturing spatiotemporal dependencies~\cite{seo2018structured,8917706}. DCRNN~\cite{li2017diffusion} modeled traffic as a diffusion process on directed graphs, using bidirectional diffusion convolution and RNNs to outperform baselines. Gated-STGCN~\cite{yu2017spatio} leveraged graph convolutions for spatial and gated 1D convolutions for temporal modeling, improving efficiency by removing recurrent structures. ASTGCN~\cite{guo2019attention} employed attention mechanisms to capture spatial and temporal dependencies, while HGCN~\cite{guo2021hierarchical} modeled traffic across micro- and macro-level networks. GWNET~\cite{wu2019graph} combined dilated causal and graph convolutions for hidden dependency capture, achieving state-of-the-art results, and OGCRNN~\cite{guo2020optimized} integrated GCN with recurrent units for enhanced feature extraction.

\subsection{Distributed and Parallel GNNs}
The growing computational demands of deep learning and the inability to deliver fast serving drove a shift from centralized to distributed computing to leverage parallelism in GNNs. Data parallelism partitions input data across nodes, using graph parallelism to minimize inter-node communication or mini-batch parallelism with node sampling techniques. Model parallelism distributes the model itself through operator parallelism with simultaneous operations, pipeline parallelism which processes stages sequentially, or ANN parallelism to spread layers across devices~\cite{besta2024parallel}.

\begin{sloppypar}
GPipe~\cite{huang2019gpipe} applied pipeline parallelism by dividing models into stages across accelerators for micro-batch processing, while PipeDream~\cite{narayanan2019pipedream} enhanced this with the 1F1B algorithm for concurrent forward and backward passes. A recent work, GNNPipe~\cite{chen2023gnnpipe} optimized GNN layer distribution across GPUs, reducing communication overhead and enabling hybrid parallelism for large scales.
\end{sloppypar}

\section{Problem Formulation} \label{sec:problemformulation}
\begin{sloppypar}
In this section, we formally model the traffic forecasting task within the framework of graph-based representation and distributed pipeline-parallel inference to define system objectives, laying the foundation for our proposed framework.
\end{sloppypar}

\subsection{Traffic Forecasting}
We model the traffic network as a graph \( G = (V, E, A) \), where \( V \) is the set of traffic observation points with \( N \) nodes, i.e., \( |V| = N \). \( A \in \mathbb{R}^{N \times N} \) defines the adjacency and \( E \) is the set of edges, where \( e_{ij} \in E \) denotes a connection between nodes \( v_i \) and \( v_j \). Each node \( v_i \) at time \( t \) has a \( d \)-dimensional feature vector \( x_i^t \in \mathbb{R}^d \). The entire traffic network at time \( t \) is represented by the feature matrix \( X_t \in \mathbb{R}^{N \times d} \), with each row corresponding to the feature vector of a node.

The traffic forecasting task aims to predict the future traffic state \( \mathcal{X}_{t+1:t+T'} = [X_{t+1}, \ldots, X_{t+T'}] \) of \( T' \) future time steps from \( t+1 \) to \( t+T' \), based on historical traffic data \( \mathcal{X}_{t-T+1:t} = [X_{t-T+1}, \ldots, X_t] \) of \( T \) time steps from \( t-T+1 \) to \( t \). The objective is to learn a nonlinear mapping \( f(\cdot) \), which captures the spatiotemporal dependencies from historical traffic observations to predict future traffic states:
\begin{equation}
\mathcal{X}_{t-T+1:t} = [X_{t-T+1}, \ldots, X_t] \xrightarrow{f(\cdot)} [X_{t+1}, \ldots, X_{t+T'}].
\end{equation}

\subsection{Pipeline-Parallel Model Inference}
Given a GNN model \( \mathcal{M} \) that consists of \( L \) layers for traffic forecasting, denoted as \( l_j \) (where \( j = 1, 2, \dots, L \)). Each layer \( l_j \) is characterized by its memory consumption \( \text{mem}_j \), input parameter size \( \text{input}_j \), and output parameter size \( \text{output}_j \). To facilitate a more streamlined inference process, we partition \( \mathcal{M} \) into multiple stages and subdivide a mini-batch of size \( B \) into multiple micro-batches, each of size \( B_{\mu} \), allowing for better resource utilization and overlap of computation and communication. Each \( \text{stage}_k \) is defined as:
\begin{equation}
\text{stage}_k = \{ l_{start_k}, l_{end_k} \}, \quad \bigcup_{k=1}^{K} \text{stage}_k = \mathcal{M}.
\end{equation}

\begin{sloppypar}
Every stage consists of a set of contiguous model layers \( l_1, l_2, \dots, l_L \), and is assigned to a single device. This process requires solving a combinatorial optimization problem to balance performance and resource constraints. Generated stages are allocated to a heterogeneous fog cluster that has the device set \( \mathcal{D} = \{ d_1, d_2, \dots, d_{Nd} \} \) containing \( N_d \) heterogeneous devices, each with distinct computational capacity, memory capacity, and communication bandwidth. The layer-device scheduling strategy \( S \) is defined as the mapping:
\begin{equation}
\begin{aligned}
S &= \{ (d_i, \text{stage}_k) \mid i, k \in \{1, 2, \dots, n\}, \; n \leq N_d; \\
&\quad \quad \quad \quad \quad \quad L_k = \{ l_{start_k}, \dots, l_{end_k} \} \}.
\end{aligned}
\end{equation}
\end{sloppypar}

Our goal is to find a joint strategy \( (\mathcal{O}, S) \), with layer execution order \( \mathcal{O} \) and layer-device scheduling strategy \( S \), that determines the partitioning of all layers of \( \mathcal{M} \) into stages and their allocation to heterogeneous fog devices for distributed parallel inference in a pipelined manner. \( (\mathcal{O}, S) \) leads to two optimization objectives.

\subsubsection{Minimizing Longest Stage Execution Time}
In pipeline parallelism, maximizing pipeline throughput is equivalent to minimizing the execution time of the longest stage.
The execution time \( T_{\text{exec}}(d_i) \) for device \( d_i \) involves computation time \( T_{\text{comp}}(\text{stage}_i, d_i) \) and intermediate value transmission time \( T_{\text{comm}}(\text{stage}_i, d_{i-1}, d_i) \):

\begin{equation}
T_{\text{exec}}(d_i) = \max \left(
\begin{aligned}
& T_{\text{comp}}(\text{stage}_i, d_i), \\
& T_{\text{comm}}(\text{stage}_i, d_{i-1}, d_i)
\end{aligned}
\right),
\end{equation}

\begin{equation}
T_{\text{comp}}(\text{stage}_i, d_i) = \sum_{l_j \in \text{stage}_i} \text{proc}_i(l_j),
\end{equation}

\begin{equation}
T_{\text{comm}}(\text{stage}_i, d_{i-1}, d_i) = \frac{\text{output}_{l_{end_i}} \cdot B_{\mu}}{\min(b_{i-1}^{\text{up}}, b_i^{\text{down}})}.
\end{equation}

\( \text{Proc}_i(l_j) \) represents the processing time for layer \( l_j \) on device \( d_i \), and the available communication bandwidth between two devices, \( b_{i-1,i} \), is bounded by the sender's up-link bandwidth \( b_{i-1}^{\text{up}} \)and the receiver's down-link bandwidth \( b_i^{\text{down}} \).

\subsubsection{Load Balancing}
To mitigate pipeline inefficiencies caused by load imbalance, which introduces idle times on faster devices, we aim to balance the load across devices by minimizing the standard deviation \( \sigma \) of devices' execution time \( T_{\text{exec}}(d_i) \):

\begin{equation}
\sigma = \sqrt{\frac{1}{N_d} \sum_{i=1}^{N_d} \left( T_{\text{exec}}(d_i) - \overline{T_{\text{exec}}} \right)^2}.
\end{equation}

\subsubsection{Optimization Summary}\sloppy
In conclusion, to maximize the pipeline throughput, we optimize both the layer execution order \( \mathcal{O} \) and layer-device scheduling \( S \), searching for the optimal \( (\mathcal{O}^*, S^*) \) that minimizes the maximum execution time across all devices:
\begin{equation}
\min_{(\mathcal{O}^*, S^*)} \max_{1 \leq i \leq N_d} \left\{
\begin{aligned}
\max \big( & T_{\text{comp}}(\text{stage}_i, d_i), \\
& T_{\text{comm}}(\text{stage}_i, d_{i-1}, d_i) \big)
\end{aligned}
\right\}. \label{optimization_objectives}
\end{equation}

\setlength{\textfloatsep}{0pt}
\begin{figure}[t]
\centerline{\includegraphics[width=7.5cm, height=6.5cm]{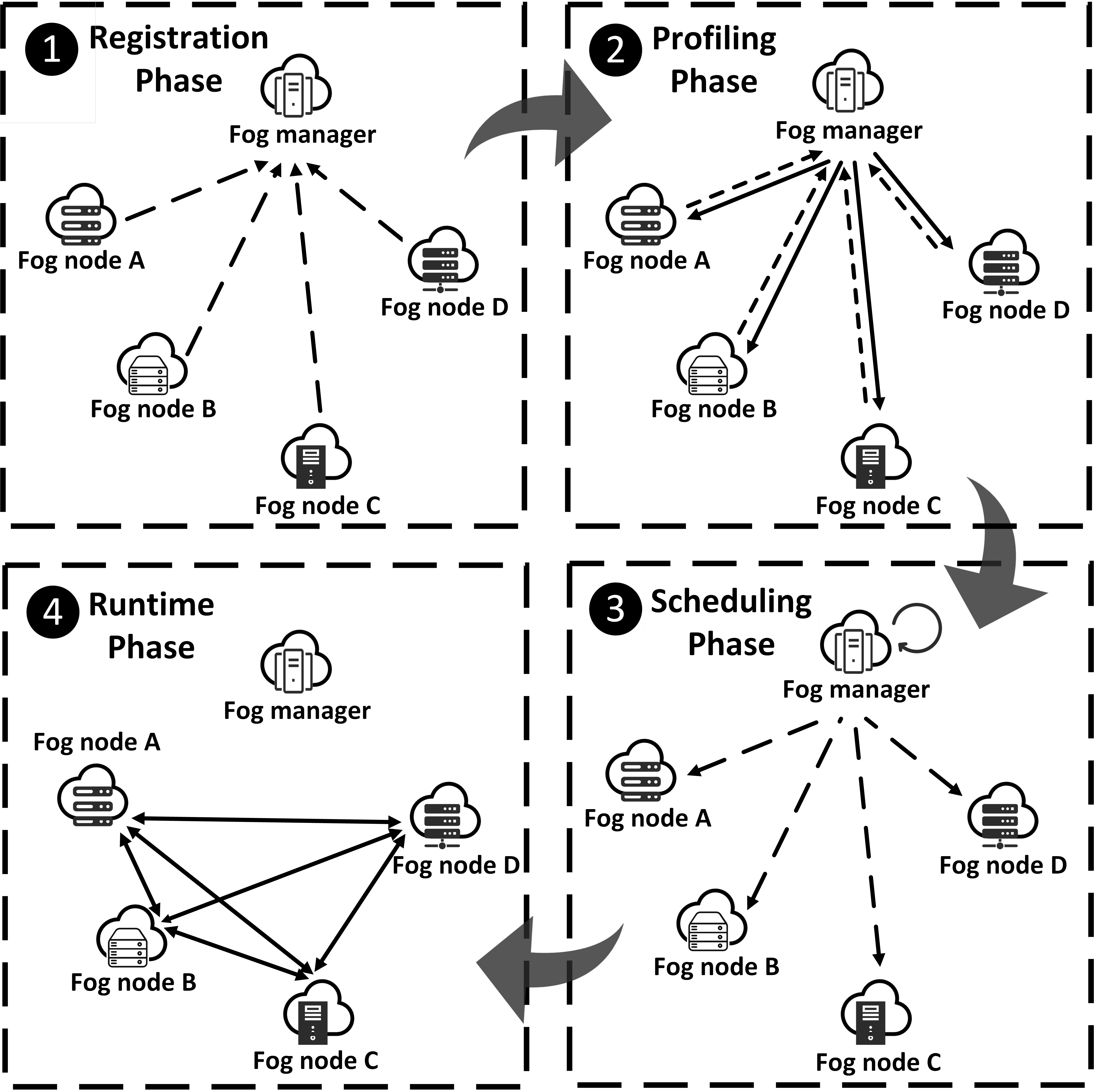}}
\caption{The overview of MultiGran-STGCNFog.}
\Description{System overall diagram.}
\label{Servers}
\end{figure}

\setlength{\textfloatsep}{0pt}
\begin{figure*}[t]
\centerline{\includegraphics[width=\linewidth, height=8.5cm]{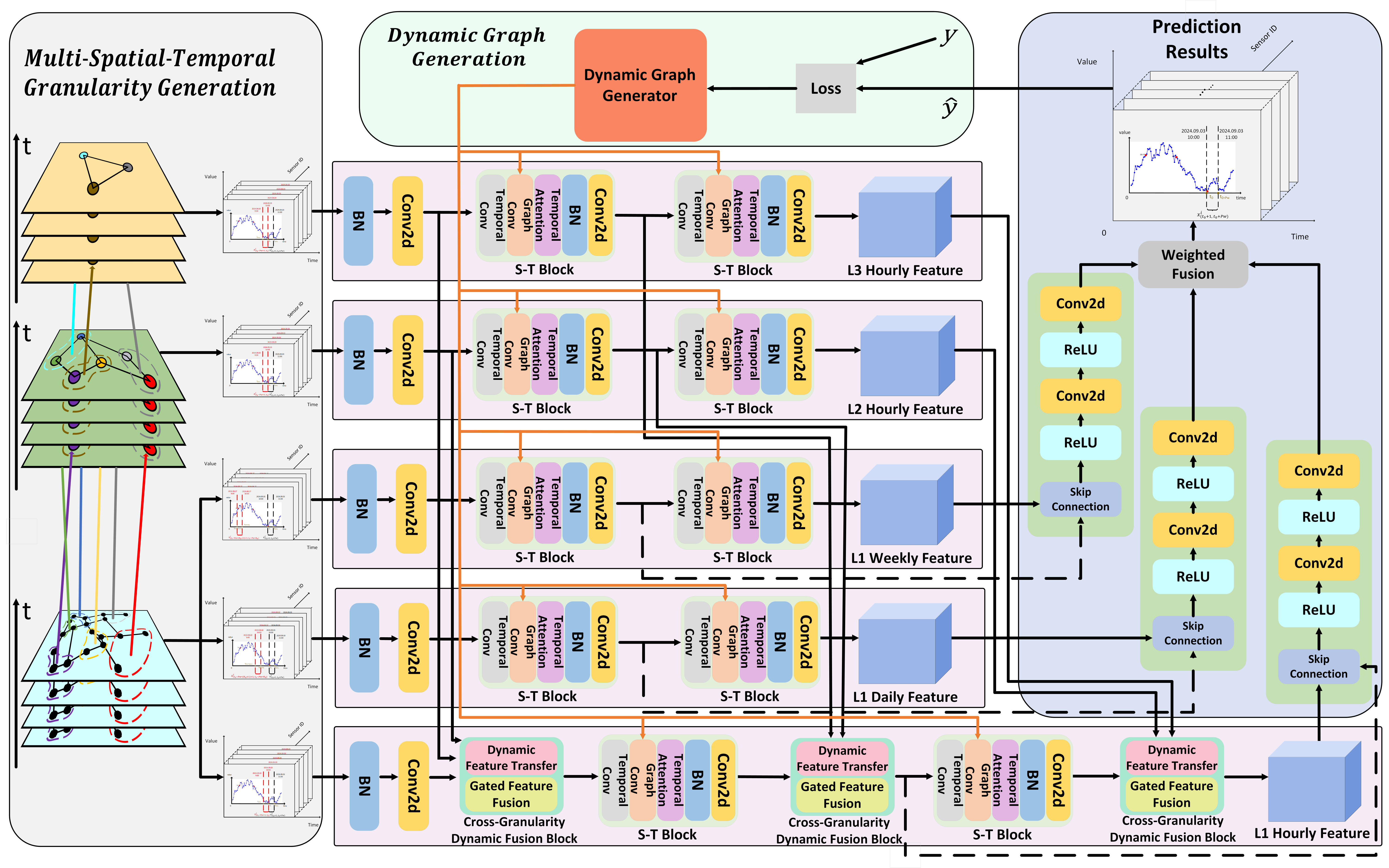}}
\caption{The network architecture of MultiGran-STGCN.}
\Description{Network architecture.}
\label{model}
\end{figure*}

\section{System Design: MultiGran-STGCNFog} \label{sec:methodology}

This section begins with a system overview illustrating all phases of the proposed inference system. Next, MultiGran-STGCN components are elaborated in detail, along with the heterogeneous cross-device execution scheduling algorithm. Lastly, a theoretical analysis establishing a lower bound on throughput improvement across varying datasets is conducted to demonstrate the robustness.

\subsection{System overview}
The workflow of MultiGran-STGCNFog consists of four phases shown in Fig.~\ref{Servers}: (1) \textbf{Registration Phase:} Fog nodes connect to the fog manager for node registration and initialization. (2) \textbf{Profiling Phase:} The fog manager distributes the forecasting model \( \mathcal{M} \) and sampled historical traffic data to registered nodes, where inference is performed to profile computation time \( \text{proc}_i(l_j) \), memory usage \( \text{mem}_j \), input and output parameter sizes, \( \text{input}_j \) \& \( \text{output}_j \) for every layer, and P2P communication bandwidth \( b_{i,i+1} \). (3) \textbf{Scheduling Phase:} The fog manager calculates the optimal scheme \( (\mathcal{O}^*, S^*) \) and assigns stage allocation with corresponding model segments. (4) \textbf{Runtime Phase:} Fog nodes perform inference collaboratively, exchanging intermediate values as needed.

\subsection{Traffic forecasting model}

The architecture of MultiGran-STGCN is depicted in Fig.~\ref{model}, and its four components will be introduced sequentially in this section.

\subsubsection{Multi-scale Traffic Data Modeling}

Three spatial scales are constructed for multi-spatial granularity generation. With the distance matrix \( D \), distances are converted into a similarity measure via a Gaussian kernel:
\begin{equation}
A(i, j) = \exp \left( -\frac{d_{ij}}{\text{var}(d_i)} \right), \quad 0 \leq A(i, j) \leq 1.
\end{equation}
The Laplacian matrix \( L \) is computed and normalized as:
\begin{equation}
L = D_{\text{deg}} - A, \quad L_{\text{norm}} = D_{\text{deg}}^{-\frac{1}{2}} L D_{\text{deg}}^{-\frac{1}{2}}.
\end{equation}

Clustering is applied based on the eigenvectors \( \textbf{H} \), which is gained by eigenvalue decomposition on \( L_{\text{norm}} \), minimizing:
\begin{equation}
\min \sum_{i=1}^{n} \sum_{j=1}^{k} \left\| \textbf{H}_i - \mu_j \right\|^2,
\end{equation}
where \( \mu_j \) is the centroid of cluster \( j \). Each cluster is treated as a node, $\mathbf{X}_i \in \mathbb{R}^{T \times d}$, in the coarser spatial scale, and features are aggregated through a series of pooling operations:
\begin{equation}
\begin{aligned}
\mathbf{X}_{\text{min}}^{(j)} &= \min_{i \in \mathcal{C}_j} \mathbf{X}_i,\;
\mathbf{X}_{\text{mean}}^{(j)} = \frac{1}{|\mathcal{C}_j|} \sum_{i \in \mathcal{C}_j} \mathbf{X}_i,\;
\mathbf{X}_{\text{max}}^{(j)} = \max_{i \in \mathcal{C}_j} \mathbf{X}_i, \\
\phantom{\mathbf{X}_{\text{min}}^{(j)}} &\quad \mathbf{X}_{\text{cluster}}^{(j)} = \text{concat}\left(\mathbf{X}_{\text{min}}^{(j)}, \mathbf{X}_{\text{mean}}^{(j)}, \mathbf{X}_{\text{max}}^{(j)}\right).
\end{aligned}
\end{equation}

\sloppypar
Three different temporal scales are set up based on original traffic data $\mathbf{X} \in \mathbb{R}^{T \times N \times d}$, namely short-term, medium-term, and long-term scales, corresponding to hourly data $\mathbf{X}_{\text{hour}} \in \mathbb{R}^{T_h \times N \times d}$, daily data $\mathbf{X}_{\text{day}} \in \mathbb{R}^{T_d \times N \times d}$, and weekly data $\mathbf{X}_{\text{week}} \in \mathbb{R}^{T_w \times N \times d}$. In detail, short-term modeling captures local dynamic changes over recent time steps. Setting the short-term window length as $W_s$, traffic data from $t_0 - h + 1$ to $t_0$ is extracted, to construct the short-term scale $\mathbf{S}_t = \left[ \mathbf{X}_{t_0 - h + 1}, \mathbf{X}_{t_0 - h + 2}, \dots, \mathbf{X}_{t_0} \right] \in \mathbb{R}^{W_s \times N \times d}$, using the short-term offset $h$. Mid-term scale captures periodic traffic patterns at daily peaks, formulated as $\mathbf{M}_t = \left[ \mathbf{X}_{t_0 - \delta_d \cdot q}, \mathbf{X}_{t_0 - \delta_d \cdot (q-1)}, \dots, \mathbf{X}_{t_0 - \delta_d} \right] \in \mathbb{R}^{W_d \times N \times d}$, where $\delta_d$ is the time offset for daily periodicity, $W_d$ is the mid-term window length. Similarly, the long-term scale $\mathbf{L}_t = \left[ \mathbf{X}_{t_0 - 7 \cdot \delta_d \cdot q}, \mathbf{X}_{t_0 - 7 \cdot \delta_d \cdot (q-1)}, \dots, \mathbf{X}_{t_0 - 7 \cdot \delta_d} \right] \in \mathbb{R}^{W_w \times N \times d}$, models trend changes over larger time scales, capturing weekly pattern from the same time points. Afterward, each temporal scale is trimmed according to the minimum number of samples across the three scales $T_{\text{min}} = \min(T_h, T_d, T_w)$.

\subsubsection{Dynamic Graph Generation Block}
Unlike traditional graph convolution operators, defined by a static graph structure \(\hat{A}\):
\begin{equation}
H^{(l+1)} = \sigma\left( \hat{D}^{-\frac{1}{2}} \hat{A} \hat{D}^{-\frac{1}{2}} H^{(l)} W^{(l)} \right).
\end{equation}
It is designed to dynamically generate the adjacency matrix \(A_t\) at each time step, allowing the graph structure to adapt to changing input data. For each node \( i \), two embedding vectors \( \mathbf{v}_{1,i} \in \mathbb{R}^d \) and \( \mathbf{v}_{2,i} \in \mathbb{R}^d \) are assigned, representing the roles of nodes as senders and receivers during the graph information dissemination. Dynamic adjacency matrix \(A_t\) is generated through node similarity calculation by the inner product:
\begin{equation}
A_{ij} = \text{ReLU}(\mathbf{v}_{1,i} \cdot \mathbf{v}_{2,j}) = \text{ReLU}\left(\sum_{k=1}^{d} v_{1,i,k} \cdot v_{2,j,k}\right).
\label{dynamic graph}
\end{equation}

\subsubsection{Feature Extraction Branches} We introduced a \textbf{Cross-Granularity Dynamic Fusion Block (C-GF block)} to further enhance the extraction of spatiotemporal correlations, which integrates feature across different scales by adjusting the fusion weights of features from different granularity levels dynamically.

The primary components of a S-T Block include \textbf{Temporal Convolution}, \textbf{Graph Convolution}, and \textbf{Temporal Attention Mechanism}. The temporal convolution extracts temporal dependencies from input data $X \in \mathbb{R}^{B \times C \times N \times T}$:
\begin{equation}
X_{\text{temp}} = \tanh(W_{\text{time}} * X + b_{\text{time}}),
\end{equation}
where $W_{\text{time}}$ is the temporal convolution kernel. To enhance feature extraction, $X_{\text{temp}}$ is then split into two parts:
\begin{equation}
X'_{\text{temp}} = \tanh(X_{\text{temp}_1}) \cdot \sigma(X_{\text{temp}_2}).
\end{equation}

The graph convolution captures spatial dependencies with the dynamically generated topology $A_{\text{dynamic}}$ by \eqref{dynamic graph} :
\begin{equation}
X_{\text{gcn}} = \tilde{D}^{-1/2} A_{\text{dynamic}} \tilde{D}^{-1/2} X'_{\text{temp}} W_{\text{gcn}}.
\end{equation}

Having $A_{\text{dynamic}}$ representing weighted spatial connections among nodes, it is similar to the temporal attention mechanism that dynamically assigns weights to different time steps by Scaled Dot-Product Attention with queries (Q), keys (K), and values (V):
\begin{equation}
\begin{split}
Q = X_{\text{gcn}} W_Q, \quad K = X_{\text{gcn}} W_K, \quad V = X_{\text{gcn}} W_V, \\
\text{Attention}(Q, K, V) = \text{softmax}\left(\frac{Q K^T}{\sqrt{d_k}}\right) V.
\end{split}
\end{equation}

With the extraction of spatiotemporal dependencies, it becomes crucial to fuse multi-granular feature representations. C-GF blocks integrate multi-granular feature representations through learnable gates that regulate the contribution of each input feature set to the final fused representation. Features from spatial scales: \( X_{L1}\), \( X_{L2}\), and \( X_{L3}\) are first concatenated to form a joint representation \(X_{\text{concat}}\). Subsequently, the gating mechanism adjusts the contribution by the learnable weight matrix \( W_g \) followed by a fusion step:
\begin{equation}
\mathbf{G} = \sigma(W_g \cdot X_{\text{concat}} + b_g),
\end{equation}
\begin{equation}
X_{\text{fused}} = \mathbf{G}_{L1} \odot X_{L1} + \mathbf{G}_{L2} \odot X_{L2} + \mathbf{G}_{L3} \odot X_{L3},
\end{equation}
\begin{equation}
X_{\text{fusion}} = \sigma(W_{\text{fusion}} \cdot X_{\text{fused}} + b_{\text{fusion}}).
\end{equation}

\subsubsection{Traffic Forecasting Head}
In the forecasting head, to enable dynamical adjustment on the contribution of each time scale based on their importance, learnable weight matrices are exploited to compute the final prediction operating the output features \( x_{\text{hour}} \), \( x_{\text{day}} \), and \( x_{\text{week}} \) from different feature extraction branches:
\begin{equation}
x = W_{\text{hour}} \odot x_{\text{hour}} + W_{\text{day}} \odot x_{\text{day}} + W_{\text{week}} \odot x_{\text{week}}.
\end{equation}



\begin{figure}[t]
\centerline{\includegraphics[width=\linewidth]{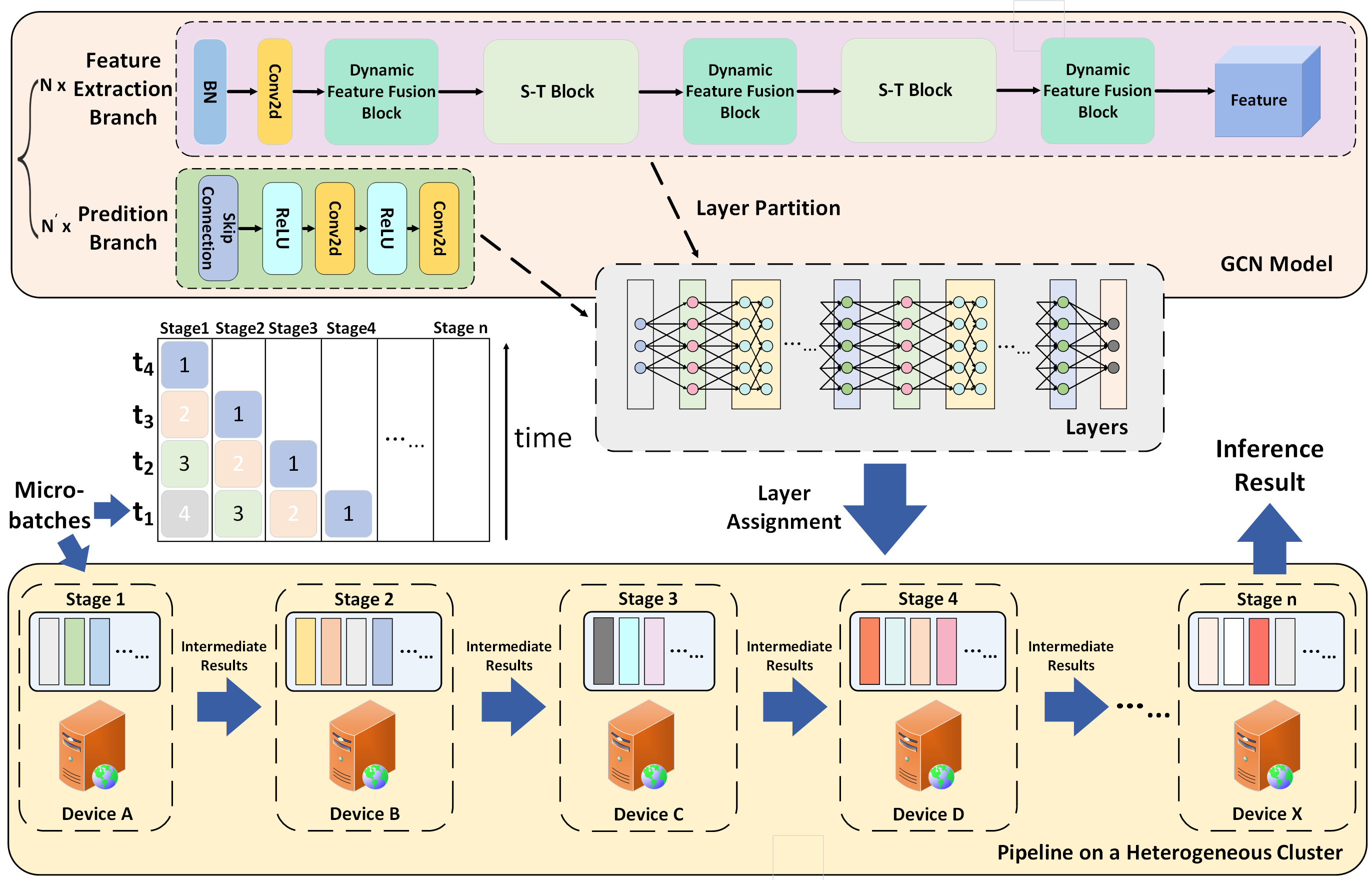}}
\caption{The illustrative diagram of inference acceleration through pipeline parallelism.}
\label{Scheduling Algorithm}
\Description{Scheduling figure}
\end{figure}

\setlength{\textfloatsep}{0pt}
\begin{algorithm}[t]
\caption{GA-DPHDS: Genetic Algorithm with Dynamic Programming for Heterogeneous Device Scheduling}
\label{alg:GA-DPHDS}
\SetKwInOut{Input}{Input}
\SetKwInOut{Output}{Output}

\Input{$L$: set of layers, $D$: set of devices, $P$: processing time matrix, $B$: bandwidth between devices, $N_{\text{pop}}$: population size, $N_{\text{gen}}$: number of generations, $P_c$: crossover probability, $P_m$: mutation probability}
\Output{$S^*$: optimal scheduling strategy}

Initialize population $\text{Pop}$ of size $N_{\text{pop}}$\;
\For{each individual $\text{ind}_i$ in Pop}{
    Decode $\text{ind}_i$ to get layer execution order $O_i$\;
    Apply DP to get scheduling strategy $S_i$\;
    Calculate $T_{\text{overall}}(\text{ind}_i)$ and $\sigma(\text{ind}_i)$\;
}
\For{$\text{gen} = 1$ to $N_{\text{gen}}$}{
    Perform fast non-dominated sorting on Pop\;
    Calculate $\text{Distance}(ind_i)$ for each $\text{ind}_i$\;
    Select parents based on rank and $\text{Distance}(ind_i)$\;
    Apply crossover and mutation with $P_c$ and $P_m$\;
    \For{each offspring $\text{ind}'_i$}{
        Decode $\text{ind}'_i$ to obtain new layer order $O'_i$\;
        Apply DP to get scheduling strategy $S'_i$\;
        Calculate $T_{\text{overall}}(\text{ind}'_i)$ and $\sigma(\text{ind}'_i)$\;
    }
    Integrate population and offspring\;
    Select top $N_{\text{pop}}$ individuals for the next generation\;
}
\Return optimal \((O^*, S^*)\) based on $T_{\text{overall}}$ and $\sigma$\;

\BlankLine
\textbf{Dynamic Programming for Scheduling Strategy:}\\
Initialize DP table $T[i][j]$ with infinity\;
Set $T[0][0] \leftarrow 0$\;
\For{$i = 1$ to $|N|$}{
    \For{$j = 1$ to $|M|$}{
        \For{$k = 0$ to $i-1$}{
            $T_{\text{comp}} \gets \sum_{l=k+1}^{i} P[l][D_j]$\;
            $T_{\text{comm}} \gets \frac{\text{output}_k \cdot B_{\mu}}{b_{j-1,j}}$\;
            $T_{\text{stage}} \gets \max(T_{\text{comp}},\ T_{\text{comm}})$\;
            Update DP table $T[i][j] \leftarrow \min\left( T[i][j],\ \max\left( T[k][j-1],\ T_{\text{stage}} \right) \right)$\;
        }
    }
}
Backtrack $T[i][j]$ to find the optimal $S^{o}$ given a $O$\;
\Return $S^{o}$ and min-max stage execution time\;
\end{algorithm}

\subsection{Cross-device execution scheduling}
Tackling the prolonged inference time due to additional feature extraction branches, we accelerate the inference process of MultiGran-STGCN through pipeline parallelism, as shown in Fig.~\ref{Scheduling Algorithm}. Specifically, the entire traffic forecasting model is partitioned into atomic unit layers and allocated to heterogeneous fog devices as stages to perform distributed pipeline-parallel inference.

Explained by \eqref{optimization_objectives}, \((O^*, S^*)\) is vital in determining how layers should be orchestrated and assigned. It is worth noting that previous works often ignored the optimization for the layer execution order \( O \). However, our experiments show that the layer execution order plays a decisive role regarding the pipeline inference throughput. We model the input-output dependencies between layers using a Directed Acyclic Graph (DAG), where each node represents a layer \(l_j\). If the output of layer \(l_j\) is required as input for layer \(l_{j+1}\), a directed edge is established from \(l_j\) to \(l_{j+1}\). This DAG structure captures the sequential and parallel execution constraints between layers. As the number of layers \(L\) increases, the search space for both \(O^*\) and \(S^*\) grows exponentially. Hence, we propose GA-DPHDS, an algorithm utilizing the NSGA-II genetic algorithm with dynamic programming for optimization, depicted by Algorithm~\ref{alg:GA-DPHDS}.

First, the population \( \text{Pop} \) is randomly initialized, where each individual \( \text{ind}_i \) represents a layer execution order \(O\), and individuals must satisfy the topological ordering constraints imposed by the DAG. For each \( \text{ind}_i \), the overall inference time \(T_{\text{overall}}(\text{ind}_i)\), longest device execution time; and the load balancing factor \(\sigma(\text{ind}_i)\), the standard deviation of execution times across devices, are calculated:

\begin{equation}
\begin{aligned}
T_{\text{overall}} = \max_{d_i \in \mathcal{D}} 
\max \big( 
&\sum_{l_j \in \text{stage}_i} T_{\text{comp}}(l_j, d_i), \\
&T_{\text{comm}}(d_{i-1}, d_i)
\big),
\end{aligned}
\end{equation}

\begin{equation}
\sigma(\text{ind}_i) = \sqrt{\frac{1}{N_d} \sum_{i=1}^{N_d} \left( T_{\text{exec}}(d_i) - \overline{T_{\text{exec}}} \right)^2}.
\end{equation}

Next, Fast Non-dominated Sorting is applied to rank the individuals based on their performance in both objective functions. We calculate the crowding distance of \( \text{ind}_i \), which quantifies how isolated an individual is in the objective space:
\begin{equation}
\begin{aligned}
\text{Distance}(\text{ind}_i) = & \, 
\frac{T_{\text{overall}}(\text{ind}_{i+1}) - T_{\text{overall}}(\text{ind}_{i-1})}{T_{\text{overall, max}} - T_{\text{overall, min}}} \\
&+ \frac{\sigma(\text{ind}_{i+1}) - \sigma(\text{ind}_{i-1})}{\sigma_{\text{max}} - \sigma_{\text{min}}}.
\end{aligned}
\end{equation}

Subsequently, selection, crossover, and mutation operations are performed. Tournament selection is based on rank and crowding distance, and crossover generates offspring under DAG constraints by exchanging segments of the parent individuals' genes. Swap mutation introduces random changes to the execution order, still adhering to DAG constraints.

For each \( \text{ind}_i \), we leverage dynamic programming to further optimize the layer-device scheduling scheme \( S \), defining a two-dimensional dynamic programming table \( T\), where \( T[i][j] \) represents the min-max execution time when assigning the first \( i \) layers to the first \( j \) devices. The state transition recurrence is given by:
\begin{equation}
T[i][j] = \min_{k < i} \left( \max \left( T[k][j-1],\ T_{\text{stage}}(k+1, i, D_j) \right) \right).
\end{equation}
This formula indicates that we need to find a splitting point \( k \) such that the layers from \( k+1 \) to \( i \) are assigned to the current device \( D_j \), while the first \( k \) layers have already been optimally assigned to devices \( D_1 \) to \( D_{j-1} \). The minimal maximum execution time \( T[i][j] \) consists of the maximum of:

1) The minimal maximum execution time of the first \( k \) layers on first \( j-1 \) devices, \( T[k][j-1] \).

2) Stage execution time on device \( D_j \), given by
\begin{equation}
T_{\text{stage}}(k+1,\ i,\ D_j) = \max \left\{
\begin{aligned}
& T_{\text{comp}}(k+1,\ i,\ D_j), \\
& T_{\text{comm}}(D_{j-1},\ D_j)
\end{aligned}
\right\}.
\end{equation}
Here, \( T_{\text{comp}}(k+1,\ i,\ D_j) \) represents the computation time for device \( D_j \) to process layers \( k+1 \) to \( i \), calculated as:
\begin{equation}
T_{\text{comp}}(k+1,\ i,\ D_j) = \sum_{l=k+1}^{i} T_{\text{comp}}(l,\ D_j),
\end{equation}
and \( T_{\text{comm}}(D_{j-1},\ D_j) \) is the communication time between devices \( D_{j-1} \) and \( D_j \), formulated as:
\begin{equation}
T_{\text{comm}}(D_{j-1}, D_j) = 
\begin{cases} 
\frac{\text{output}_k \cdot B_\mu}{b_{j-1, j}}, & \text{if } D_{j-1} \neq D_j, \\
0, & \text{if } D_{j-1} = D_j.
\end{cases}
\end{equation}

With such a state transition, the dynamic programming table \( T\) is updated iteratively to reflect the optimal solution \((O^*, S^*)\) to the cross-device execution scheduling.

\subsection{Theoretical Guarantee of Throughput Improvement under Dataset Variations}\label{sec:theory-throughput}

While input variations rarely impact throughput in standard inference with fixed model structure and hardware, distributed GNN inference may exhibit performance shifts due to changes in graph structure induced by the dataset. To address this, we derive a theoretical lower bound on throughput improvement under dataset variations, providing formal efficiency guarantees of GA-DPHDS with respect to the indirect impact of datasets.

\textbf{Problem Setup:}  
We consider a fixed model \( \mathcal{M} \) consisting of $L$ layers $\{l_1, l_2, \dots, l_L\}$ deployed over a set of $K$ devices $\mathcal{D} = \{d_1, \dots, d_K\}$. The micro-batch size $B_\mu$, network bandwidths $b_{i-1,i}$, and model architecture are fixed. The only varying factor is the input dataset $D \in \mathcal{D}_{\text{data}}$, which affects the runtime profile of each layer. We define:
\begin{itemize}
  \item $t_j^D$: compute time of layer $l_j$ under dataset $D$.
  \item $o_j^D$: output tensor size of layer $l_j$.
  \item $o_{\max}^D = \max_j o_j^D$; $o_{\text{avg}}^D = \frac{1}{L} \sum_j o_j^D$: maximum and average output tensor size.
  \item $\alpha_D := \frac{o_{\max}^D}{o_{\text{avg}}^D}$: output tensor imbalance ratio.
  \item $b^{\text{up}}_k$, $b^{\text{down}}_k$: uplink/downlink bandwidth of device $d_k$.
  \item $\beta := \frac{\min(b^{\text{good}})}{\min(b^{\text{bad}})}$: link bandwidth asymmetry ratio.
  \item $\delta$: the small residual imbalance that remains after scheduling, due to layer indivisibility and hardware heterogeneity.
\end{itemize}

\textbf{Pipeline Stage Time Model:}  
For any stage $S_k$ allocated to device $d_k$, the execution time is:
\begin{equation}
T_k^D = \max\left( \sum_{j \in S_k} t_j^D,\quad \frac{o_{j_{\text{last}}}^D \cdot B_\mu}{\min(b^{\text{up}}_{k-1}, b^{\text{down}}_k)} \right),
\end{equation}
where $j_{\text{last}}$ is the final layer in $S_k$.

\textbf{Throughput Definition:}  
Let $(O, S)$ be a layer ordering and device assignment:
\begin{equation}
T_{\text{max}}^D(O, S) = \max_k T_k^D, \text{Throughput}(O, S; D) = \frac{1}{T_{\text{max}}^D(O, S)}.
\end{equation}

We compare two scheduling strategies:
\begin{itemize}
  \item Baseline: $(O_0, S_0)$ (e.g., equal layer partitioning)
  \item GA-DPHDS: $(O^*, S^*)$ (optimized via genetic search)
\end{itemize}

Define the throughput improvement ratio:
\begin{equation}
\gamma_D := \frac{\text{Throughput}(O^*, S^*; D)}{\text{Throughput}(O_0, S_0; D)} = \frac{T_{\text{max}}^D(O_0, S_0)}{T_{\text{max}}^D(O^*, S^*)}.
\end{equation}

\textbf{Step 1: Baseline Bottleneck Estimation}  
In baseline scheduling, stage boundaries are fixed and may split at layers with large output tensors. If such a split falls on a poor bandwidth link, the worst-case stage time becomes:
\begin{equation}
T_{\text{baseline}}(D) \geq \frac{o_{\max}^D \cdot B_\mu}{\min(b^{\text{bad}})},
\end{equation}
which can dominate stage runtime if the split is poorly chosen.

\textbf{Step 2: GA-DPHDS Improvement}  
GA-DPHDS explicitly searches to avoid such unfavorable splits. Assuming it avoids high-output layers and selects fast links, its worst stage time is upper bounded by:
\begin{equation}
T_{\text{GA}}(D) \leq \frac{o_{\text{avg}}^D \cdot B_\mu}{\min(b^{\text{good}})} + \delta.
\end{equation}

\textbf{Step 3: Lower Bound on Improvement Ratio}  
Define $\alpha_D := \frac{o_{\max}^D}{o_{\text{avg}}^D}$ and $\beta := \frac{\min(b^{\text{good}})}{\min(b^{\text{bad}})}$, then:
\begin{equation}
\gamma_D \geq \frac{o_{\max}^D / \min(b^{\text{bad}})}{o_{\text{avg}}^D / \min(b^{\text{good}}) + \delta / B_\mu} 
= \frac{\alpha_D \cdot \beta}{1 + \frac{\delta}{B_\mu \cdot o_{\text{avg}}^D / \min(b^{\text{good}})}}.
\end{equation}
This gives an explicit lower bound on the throughput improvement, driven by $\alpha_D$, $\beta$, and $\delta$.

\begin{theorem}[Communication-Aware Throughput Gain]
Let model $M$, device cluster $\mathcal{D}$, and micro-batch size $B_\mu$ be fixed. For any graph-structured traffic data (dataset) $D$ where $\alpha_D$ is bounded, $\beta > 1$, and $\delta$ is small enough, then GA-DPHDS scheduling achieves throughput improvement:
\begin{equation}
\gamma_D \geq \frac{\alpha_D \cdot \beta}{1 + \varepsilon}, \quad \text{where } \varepsilon = \frac{\delta}{B_\mu \cdot o_{\text{avg}}^D / \min(b^{\text{good}})} \ll 1.
\end{equation}
\end{theorem}

\textbf{Practical Significance: Empirical Estimation}  
In our experimental setup shown in Table \ref{tab:device_config}, device bandwidths range from 1.2–3.2 Gbps, i.e., 150–400 MB/s.  
A realistic range for links yields:
\[
\min(b^{\text{bad}}) = 150\text{MB/s}, \ \min(b^{\text{good}}) = 200\text{MB/s} \Rightarrow \beta \approx 1.33.
\]

In our experiments, $o_{\text{avg}}^D = 1.11$MB, $o_{\max}^D = 3.60$MB $\Rightarrow \alpha_D = 3.24$, and $B_\mu = 16$.

Estimated communication time under GA:
\begin{equation}
\frac{o_{\text{avg}}^D}{\min(b^{\text{good}})} = \frac{1.11}{200} = 0.00555\text{ s}.
\end{equation}

Assume $\delta = 10$ms $= 0.005$ (converted to seconds), then:
\begin{equation}
\gamma_D \gtrapprox \frac{3.24 \cdot 1.33}{1 + 0.01 / 0.00555} = \frac{4.309}{2.802} \approx 1.54.
\end{equation}

\textbf{\textit{Conclusion:}}  
GA-DPHDS can theoretically yield \textbf{at least 1.54×} throughput improvement under realistic profile variation and bandwidth heterogeneity, which aligns well with the experimental gain \textbf{2.19×} of the cluster 4 reported in Section~\ref{subsec:HeterogeneousvsHomogeneous}.

\textit{This confirms that the theoretical advantage of GA-DPHDS remains practically significant under real hardware constraints.}

\section{Experiments} \label{sec:experiment result}

The experiments section aims to: \textbf{(1) Validate the performance of MultiGran-STGCN. (2) Demonstrate the necessity of GA-DPHDS against homogeneous counterparts. (3) Explore the impact of device network bandwidth on overall inference throughput and identify system bottlenecks. (4) Evaluate the effectiveness of our optimization approach.}

\begin{table}[htbp]
\centering
\caption{Heterogeneous device table}
\label{tab:device_config}
\renewcommand{\arraystretch}{1.0}
\Huge
\resizebox{\columnwidth}{!}{%
\begin{tabular}{cccc}
\toprule
\textbf{Device} & \textbf{CPU} & \textbf{Avail. Mem. (GB)} & \textbf{Bandwidth (Gbps)} \\
\midrule
A & i7-12700F @ 2.10 GHz & 32 & 3.2 \\
B & Xeon E3-1230 v6 @ 3.50 GHz & 16 & 2.4 \\
C & i7-9700K @ 3.60 GHz & 48 & 1.6 \\
D & M3 Pro (11-core) & 18 & 2.4 \\
E & i7-7700 @ 3.60 GHz & 16 & 2.0 \\
F & i7-9750H @ 2.60 GHz & 8 & 1.2 \\
\bottomrule
\end{tabular}%
}
\end{table}

\begin{table*}[ht]
\Huge
\centering
\caption{Performance on PEMS04, PEMS07, and PEMS08 datasets, {\textbf{Pink}}/{\textbf{Green}}/{\textbf{Beige}} marks the best/second-best/third-best performance. Avg.$\Delta$(T3) denotes the average performance improvement of MultiGran-STGCN over top-3 baselines}
\label{tab:performance_comparison}
\renewcommand{\arraystretch}{1.2}
\resizebox{\textwidth}{!}{
\begin{tabular}{llcccccccccccccc}
\toprule
 & \multicolumn{1}{r}{Metric} & HA & LSTM & GRU & GCRN & Gated-STGCN & GWNET & OGCRNN & HGCN & ASTGCN & MG-single & MG-wms & MG-wmt & MultiGran-STGCN & Avg.$\Delta$(T3) \\
\midrule

\multirow{9}{*}{\textbf{PEMS04}} 
& \multicolumn{1}{r}{MAE} 
   & 28.38 & 20.26 & 20.18 & 20.78 & 20.36 & \cellcolor{green}{18.12} & 19.76 & \cellcolor{Beige}{18.29} & 20.35 & 19.00 & 18.96 & 18.33 & \cellcolor{pink}{\textbf{18.11}} & \cellcolor{ImproveColor}{3.28\%} \\
\addlinespace[0.1em]
   
& \multicolumn{1}{r}{15 mins MAPE} 
   & 19.99\% & 13.87\% & 13.56\% & 15.78\% & 15.77\% & \cellcolor{green}{12.58\%} & 13.93\% & 13.02\% & 14.23\% & 13.74\% & 13.56\% & \cellcolor{Beige}{12.83\%} & \cellcolor{pink}{\textbf{12.45\%}} & \cellcolor{ImproveColor}{5.37\%} \\
\addlinespace[0.1em]
   
& \multicolumn{1}{r}{RMSE} 
   & 41.82 & 31.71 & 31.61 & 31.67 & 31.14 & \cellcolor{pink}{\textbf{28.83}} & 30.49 & 29.84 & 31.92 & 29.92 & 29.74 & \cellcolor{Beige}{29.08} & \cellcolor{green}{28.84} & \cellcolor{ImproveColor}{2.96\%} \\
\cmidrule{2-16}

& \multicolumn{1}{r}{MAE} 
   & 31.77 & 22.31 & 22.23 & 22.14 & 21.97 & \cellcolor{green}{18.86} & 20.47 & 19.10 & 20.93 & 19.84 & 19.80 & \cellcolor{Beige}{19.04} & \cellcolor{pink}{\textbf{18.66}} & \cellcolor{ImproveColor}{4.19\%} \\
\addlinespace[0.1em]
& \multicolumn{1}{r}{30 mins MAPE} 
   & 22.65\% & 15.39\% & 14.94\% & 16.68\% & 16.79\% & \cellcolor{green}{13.13\%} & 14.42\% & 13.67\% & 14.57\% & 14.38\% & 14.15\% & \cellcolor{Beige}{13.21\%} & \cellcolor{pink}{\textbf{12.83\%}} & \cellcolor{ImproveColor}{6.62\%} \\
\addlinespace[0.1em]
& \multicolumn{1}{r}{RMSE} 
   & 46.49 & 34.54 & 34.46 & 33.65 & 33.44 & \cellcolor{green}{29.93} & 31.55 & \cellcolor{Beige}{30.06} & 33.00 & 31.11 & 30.90 & 30.16 & \cellcolor{pink}{\textbf{29.82}} & \cellcolor{ImproveColor}{2.27\%} \\
\cmidrule{2-16}

& \multicolumn{1}{r}{MAE} 
   & 38.51 & 26.41 & 26.33 & 24.98 & 25.17 & \cellcolor{green}{20.06} & 21.74 & 20.62 & 21.87 & 21.39 & 20.89 & \cellcolor{Beige}{20.19} & \cellcolor{pink}{\textbf{19.55}} & \cellcolor{ImproveColor}{6.04\%} \\
\addlinespace[0.1em]
& \multicolumn{1}{r}{60 mins MAPE} 
   & 28.20\% & 18.65\% & 17.93\% & 18.68\% & 19.05\% & \cellcolor{Beige}{13.98}\% & 15.41\% & 14.86\% & 15.10\% & 15.47\% & 14.81\% & \cellcolor{green}{13.92\%} & \cellcolor{pink}{\textbf{13.45\%}} & \cellcolor{ImproveColor}{8.17\%} \\
\addlinespace[0.1em]
& \multicolumn{1}{r}{RMSE} 
   & 55.76 & 40.07 & 40.08 & 37.69 & 37.87 & \cellcolor{green}{31.62} & 33.37 & 32.20 & 34.82 & 33.23 & 32.77 & \cellcolor{Beige}{31.89} & \cellcolor{pink}{\textbf{31.31}} & \cellcolor{ImproveColor}{3.35\%} \\
\midrule

\multirow{9}{*}{\textbf{PEMS07}} 
& \multicolumn{1}{r}{MAE} 
   & 32.82 & 21.79 & 21.83 & 22.71 & 22.07 & 19.02 & 20.50 & 19.18 & 22.76 & 20.87 & \cellcolor{green}{18.65} & \cellcolor{Beige}{18.88} & \cellcolor{pink}{\textbf{18.56}} & \cellcolor{ImproveColor}{5.14\%} \\
\addlinespace[0.1em]
& \multicolumn{1}{r}{15 mins MAPE} 
   & 15.04\% & 9.26\% & 9.23\% & 11.59\% & 11.57\% & \cellcolor{Beige}{8.20\%} & 9.10\% & 8.29\% & 9.96\% & 11.29\% & \cellcolor{green}{8.11\%} & 8.25\% & \cellcolor{pink}{\textbf{7.99\%}} & \cellcolor{ImproveColor}{6.33\%} \\
\addlinespace[0.1em]
& \multicolumn{1}{r}{RMSE} 
   & 47.97 & 33.84 & 33.76 & 34.16 & 32.84 & 30.14 & 31.85 & 30.25 & 35.42 & 31.46 & \cellcolor{pink}{\textbf{29.72}} & \cellcolor{Beige}{29.83} & \cellcolor{green}{29.82} & \cellcolor{ImproveColor}{3.01\%} \\
\cmidrule{2-16}

& \multicolumn{1}{r}{MAE} 
   & 37.03 & 24.42 & 24.50 & 24.51 & 24.14 & 20.32 & 21.54 & 20.46 & 23.89 & 22.33 & \cellcolor{green}{19.67} & \cellcolor{Beige}{20.14} & \cellcolor{pink}{\textbf{19.52}} & \cellcolor{ImproveColor}{6.03\%} \\
\addlinespace[0.1em]
& \multicolumn{1}{r}{30 mins MAPE} 
   & 17.19\% & 10.39\% & 10.36\% & 12.17\% & 12.40\% & \cellcolor{Beige}{8.67\%} & 9.50\% & 8.77\% & 10.30\% & 12.05\% & \cellcolor{green}{8.51\%} & 8.73\% & \cellcolor{pink}{\textbf{8.38\%}} & \cellcolor{ImproveColor}{6.68\%} \\
\addlinespace[0.1em]
& \multicolumn{1}{r}{RMSE} 
   & 53.99 & 37.56 & 37.52 & 36.83 & 35.95 & 32.16 & 33.48 & 32.27 & 37.93 & 33.71 & \cellcolor{green}{31.60} & \cellcolor{Beige}{31.79} & \cellcolor{pink}{\textbf{31.55}} & \cellcolor{ImproveColor}{3.33\%} \\
\cmidrule{2-16}

& \multicolumn{1}{r}{MAE} 
   & 45.33 & 29.51 & 29.61 & 28.00 & 28.08 & 22.37 & 23.16 & 22.31 & 25.32 & 24.35 & \cellcolor{green}{20.90} & \cellcolor{Beige}{21.84} & \cellcolor{pink}{\textbf{20.76}} & \cellcolor{ImproveColor}{8.20\%} \\
\addlinespace[0.1em]
& \multicolumn{1}{r}{60 mins MAPE} 
   & 21.56\% & 12.80\% & 12.67\% & 13.45\% & 14.12\% & \cellcolor{Beige}{9.42\%} & 10.20\% & 9.54\% & 10.78\% & 13.29\% & \cellcolor{green}{9.03\%} & 9.49\% & \cellcolor{pink}{\textbf{8.95\%}} & \cellcolor{ImproveColor}{7.92\%} \\
\addlinespace[0.1em]
& \multicolumn{1}{r}{RMSE} 
   & 65.74 & 44.59 & 44.42 & 41.75 & 41.45 & 35.12 & 35.19 & 35.06 & 41.12 & 36.58 & \cellcolor{green}{34.00} & \cellcolor{Beige}{34.47} & \cellcolor{pink}{\textbf{33.81}} & \cellcolor{ImproveColor}{3.74\%} \\
\midrule

\multirow{9}{*}{\textbf{PEMS08}} 
& \multicolumn{1}{r}{MAE} 
   & 23.11 & 16.13 & 16.08 & 16.85 & 16.97 & 14.11 & 16.04 & 14.37 & 16.53 & 15.14 & \cellcolor{Beige}{14.80} & \cellcolor{green}{14.28} & \cellcolor{pink}{\textbf{14.03}} & \cellcolor{ImproveColor}{5.46\%} \\
\addlinespace[0.1em]
& \multicolumn{1}{r}{15 mins MAPE} 
   & 14.46\% & 10.02\% & 10.03\% & 11.94\% & 13.58\% & 9.47\% & 10.66\% & \cellcolor{pink}{\textbf{9.20\%}} & 10.62\% & 10.43\% & 10.37\% & \cellcolor{green}{9.24\%} & \cellcolor{Beige}{9.34\%} & \cellcolor{ImproveColor}{2.34\%} \\
\addlinespace[0.1em]
& \multicolumn{1}{r}{RMSE} 
   & 34.13 & 24.95 & 24.89 & 25.20 & 24.92 & \cellcolor{pink}{\textbf{21.65}} & 24.43 & \cellcolor{Beige}{21.92} & 25.24 & 22.78 & 22.59 & 21.95 & \cellcolor{green}{21.90} & \cellcolor{ImproveColor}{3.38\%} \\
\cmidrule{2-16}

& \multicolumn{1}{r}{MAE} 
   & 26.08 & 17.94 & 17.86 & 17.59 & 17.99 & \cellcolor{green}{14.84} & 16.77 & 15.23 & 17.01 & 16.10 & 15.37 & \cellcolor{Beige}{15.05} & \cellcolor{pink}{\textbf{14.49}} & \cellcolor{ImproveColor}{7.19\%} \\
\addlinespace[0.1em]
& \multicolumn{1}{r}{30 mins MAPE} 
   & 16.36\% & 11.06\% & 11.11\% & 12.27\% & 14.14\% & 9.84\% & 11.00\% & \cellcolor{Beige}{9.62\%} & 10.89\% & 11.09\% & 10.80\% & \cellcolor{green}{9.61\%} & \cellcolor{pink}{\textbf{9.55\%}} & \cellcolor{ImproveColor}{5.60\%} \\
\addlinespace[0.1em]
& \multicolumn{1}{r}{RMSE} 
   & 38.31 & 27.76 & 27.66 & 26.49 & 26.80 & \cellcolor{green}{22.94} & 25.68 & 23.35 & 26.27 & 24.34 & 23.73 & \cellcolor{Beige}{23.32} & \cellcolor{pink}{\textbf{22.90}} & \cellcolor{ImproveColor}{4.54\%} \\
\cmidrule{2-16}

& \multicolumn{1}{r}{MAE} 
   & 32.00 & 21.46 & 21.38 & 19.12 & 20.14 & \cellcolor{green}{16.02} & 18.17 & 16.69 & 17.61 & 17.49 & \cellcolor{Beige}{16.20} & 16.28 & \cellcolor{pink}{\textbf{15.12}} & \cellcolor{ImproveColor}{9.86\%} \\
\addlinespace[0.1em]
& \multicolumn{1}{r}{60 mins MAPE} 
   & 20.28\% & 13.25\% & 13.47\% & 13.07\% & 15.27\% & 10.41\% & 11.79\% & \cellcolor{Beige}{10.38\%} & 11.26\% & 12.13\% & 11.43\% & \cellcolor{green}{10.24\%} & \cellcolor{pink}{\textbf{9.91\%}} & \cellcolor{ImproveColor}{7.24\%} \\
\addlinespace[0.1em]
& \multicolumn{1}{r}{RMSE} 
   & 46.50 & 32.79 & 32.73 & 29.97 & 30.31 & \cellcolor{green}{24.88} & 27.87 & 25.58 & 27.56 & 26.48 & \cellcolor{Beige}{25.35} & 25.39 & \cellcolor{pink}{\textbf{24.18}} & \cellcolor{ImproveColor}{7.02\%} \\
\bottomrule
\end{tabular}
}
\end{table*}

\subsection{Experimental Setup}

\begin{table}[htbp]
\centering
\caption{Heterogeneous clusters configuration}
\label{tab:device_matrix}
\renewcommand{\arraystretch}{0.5}
\LARGE
\resizebox{\columnwidth}{!}{%
\begin{tabular}{lcccccccccc}
\toprule
\textbf{Device} & \multicolumn{10}{c}{\textbf{Cluster}} \\
\cmidrule(lr){2-11}
 & \textbf{One} & \textbf{Two} & \textbf{Three} & \textbf{Four} & \textbf{Five} & \textbf{Six} & \textbf{Seven} & \textbf{Eight} & \textbf{Nine} & \textbf{Ten} \\
\midrule
Device A & \checkmark & \checkmark & \checkmark & \checkmark &  & \checkmark &  &  & \checkmark & \checkmark \\
Device B & \checkmark & \checkmark & \checkmark & \checkmark &  &  & \checkmark & \checkmark &  & \checkmark \\
Device C & \checkmark & \checkmark & \checkmark & \checkmark &  &  & \checkmark & \checkmark &  &  \\
Device D &  & \checkmark & \checkmark & \checkmark & \checkmark & \checkmark &  &  & \checkmark &  \\
Device E &  &  & \checkmark & \checkmark & \checkmark &  & \checkmark &  &  &  \\
Device F &  &  &  & \checkmark & \checkmark & \checkmark & \checkmark &  &  &  \\
\bottomrule
\end{tabular}%
}
\end{table}

We conducted experiments using six heterogeneous fog devices with varying capabilities regarding the CPU, memory, and communication bandwidth. Table \ref{tab:device_config} provides an overview of used devices, which were configured into ten clusters with distinct memberships shown in Table \ref{tab:device_matrix}, where network communication was enabled through a symmetric local Ethernet. Performance was evaluated in terms of forecasting accuracy and inference throughput.

Three benchmark datasets, PEMS04, PEMS07, and PEMS08 were utilized to evaluate MultiGran-STGCN, compared against a set of statistical and GCN approaches introduced in Section~\ref{sec:rework}, including \textbf{HA}, \textbf{LSTM}~\cite{cui2018deep}, \textbf{GRU}~\cite{agarap2018neural}, \textbf{GCRN}~\cite{seo2018structured}, \textbf{Gated-STGCN}~\cite{yu2017spatio}, \textbf{GWNET}~\cite{wu2019graph}, \textbf{OGCRNN}~\cite{guo2020optimized}, \textbf{HGCN}~\cite{guo2021hierarchical}, and \textbf{ASTGCN}~\cite{guo2019attention}. Baselines were configured as reported in their respective works.

We leveraged \textbf{MAE}, \textbf{MAPE}, and \textbf{RMSE} to measure forecasting performance. Since the throughput gain was analytically guaranteed in Section~\ref{sec:theory-throughput}, inference efficiency was assessed via pipeline throughput on PEMS04, with network latency randomly perturbed between 10 \textasciitilde\ 30 ms to better simulate real-world fog environments.

\subsection{Spatiotemporal Traffic Forecasting}

Table \ref{tab:performance_comparison} presents the performance of MultiGran-STGCN, baselines, and ablation variants: MG-single (single spatial and temporal granularity), MG-wms (without multi-spatial granularity), and MG-wmt (without multi-temporal granularity). MultiGran-STGCN consistently outperforms baselines and its variants: across 15, 30, and 60-minute horizons, with up to 9.86\% boost over top-3 baselines.

For 15-minute forecasts, MultiGran-STGCN achieves MAE values of 18.11, 18.56, and 14.03 on PEMS04, PEMS07, and PEMS08, respectively, surpassing all baselines. It also delivers optimal or near-optimal MAPE and RMSE. For 30-minute forecasts, MultiGran-STGCN continues to lead, notably achieving the best performance across every metric on all datasets and fully surpassing any minor discrepancies observed in the short-term forecasts. This illustrates that integrating multi-granular spatiotemporal modeling and designed dynamic mechanisms further enhances MultiGran-STGCN's capability to capture complex traffic dependencies. For 60-minute forecasts, The MAE values for PEMS04, PEMS07, and PEMS08 are 19.55, 20.76, and 15.12, suggesting more significant performance advances. Furthermore, the ablation variants -wms and -wmt also show notable superiority over baselines. On PEMS07, -wms and -wmt achieve MAE, MAPE, and RMSE of 20.90, 9.03\%, and 34.00, as well as 21.84, 9.49\%, and 34.47, outpacing the best-performing baseline's metric values of GWNET.

\subsubsection{Ablation Study of Multi-spatiotemporal Scale Modeling}

\begin{figure}[ht]
\centerline{\includegraphics[width=8cm]{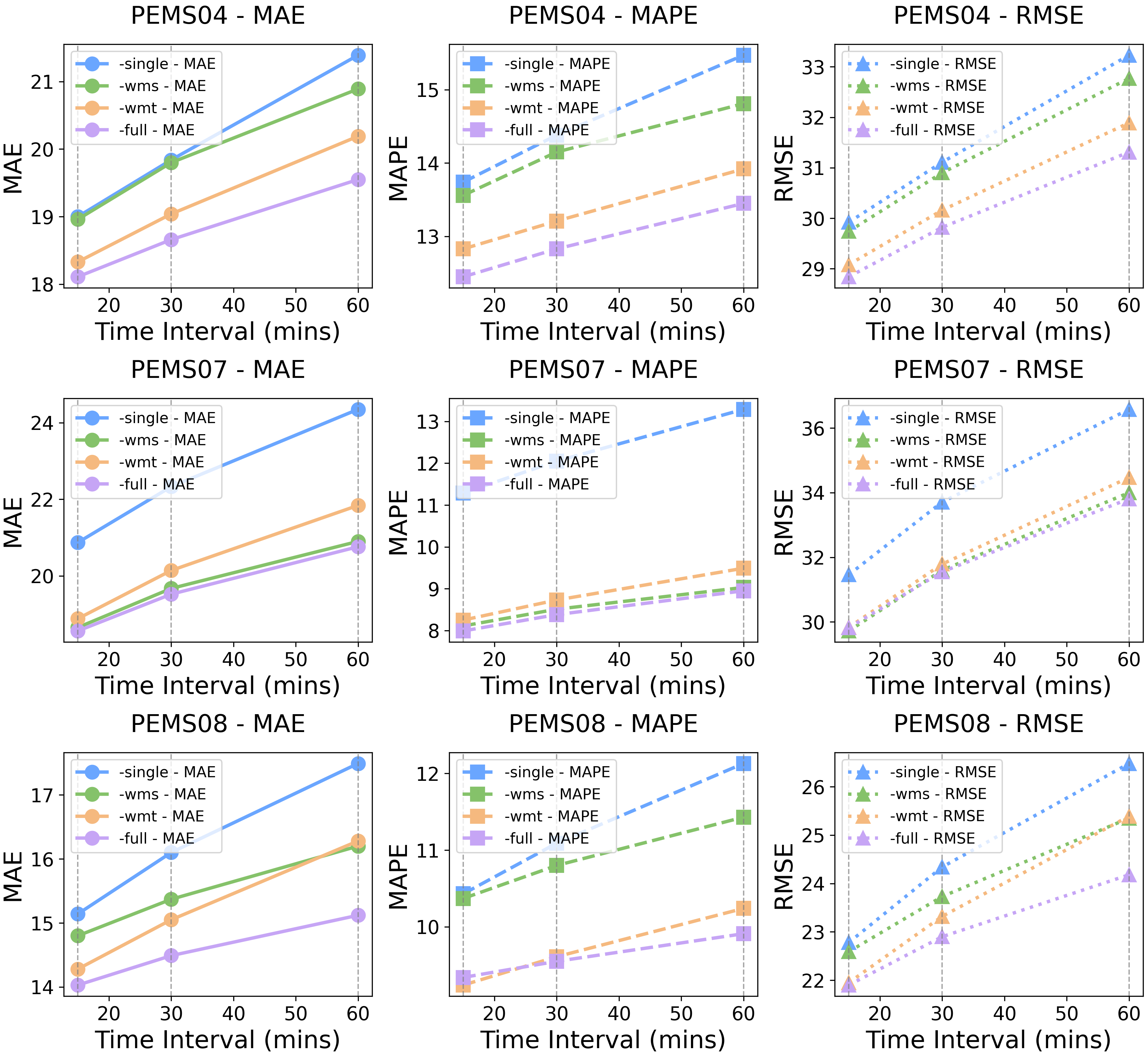}}
\caption{Ablation study: forecasts across horizons.}
\label{ablation}
\Description{Ablation}
\end{figure}

Fig.~\ref{ablation} provides a more intuitive performance comparison among ablation variants and the complete MultiGran-STGCN model denoted as -full, to explore the performance gain of multi-scale modeling and multi-granular feature fusion design. On all three datasets, the -full model consistently outperforms other variants, and its relatively modest increase in errors indicates its effectiveness in capturing long-term dependencies, as the prediction horizon extends from 15 to 60 minutes. The enormous gap between the -single variant and other models points out the limitation of relying solely on single-dimensional scales, drawbacks of previous studies. This observation suggests that while -wms and -wmt partially incorporate multi-spatiotemporal feature extraction, the lack of comprehensive integration restricts their forecasting accuracy. Besides, it is also clear that as the horizon increases, the trajectories begin to diverge. The growing divergence underscores our model's superiority in capturing long-term correlations and handling error accumulation.

\subsection{Distributed pipeline-parallel inference}

\setlength{\textfloatsep}{0pt}
\begin{figure}[htbp]
\centerline{\includegraphics[width=\linewidth, height=3.5cm]{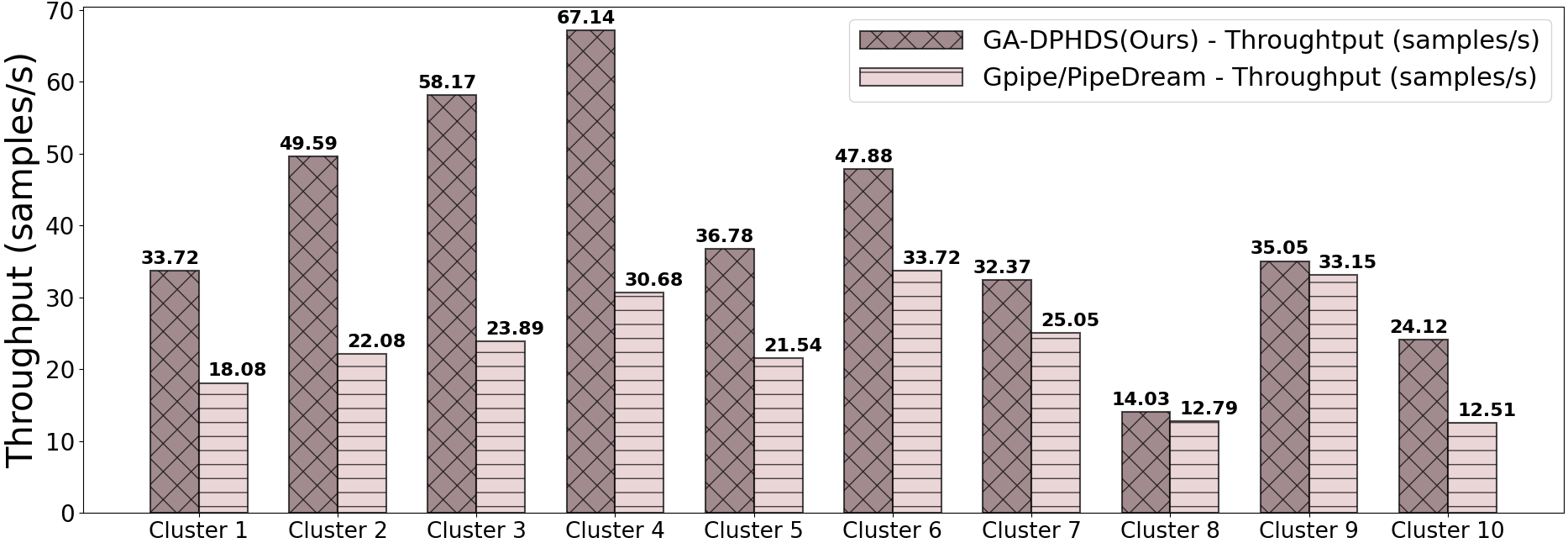}}
\caption{Heterogeneous vs. homogeneous scheduling.}
\label{heterogeneous_homogeneous}
\Description{heterogeneous_homogeneous}
\end{figure}

\subsubsection{Heterogeneous vs. Homogeneous Scheduling}\label{subsec:HeterogeneousvsHomogeneous}
The scheduling methods of GPipe~\cite{huang2019gpipe} and PipeDream~\cite{narayanan2019pipedream} were adopted as the baseline against GA-DPHDS in terms of pipeline throughput. GPipe utilized homogeneous accelerators, while PipeDream partitioned DNN operators based on a single GPU's capabilities. Devices and pipeline sequences were randomized, and each cluster was tested ten times to obtain the average. As shown in fig.~\ref{heterogeneous_homogeneous}, GA-DPHDS outperformed the baseline across all cluster configurations, achieving throughput from 14.03 samples/s (Cluster 8) to 67.14 samples/s (Cluster 4), while the baseline peaked at 33.72 samples/s and dropped to 12.51 samples/s (Cluster 10). In Cluster 4, GA-DPHDS reached 67.14 samples/s, more than doubling the baseline’s 30.68 samples/s, highlighting its efficiency in resource-intensive setups. In Cluster 2, GA-DPHDS improved throughput to 49.59 samples/s compared to 22.08 samples/s of the baseline. However, in Cluster 8, having only devices B and C, GA-DPHDS exhibited only minor improvements due to comparable device capabilities, consistent with table \ref{tab:device_throughput}. These findings underscore GA-DPHDS's effectiveness in optimizing throughput across varied cluster setups.

\begin{table}[htbp]
    \centering
    \caption{Single device throughput (samples/s)}
    \renewcommand{\arraystretch}{0.5}
    \label{tab:device_throughput}
    \resizebox{\linewidth}{!}{%
    \begin{tabular}{lcccccc}
        \toprule
        \textbf{Device} & \textbf{A} & \textbf{B} & \textbf{C} & \textbf{D} & \textbf{E} & \textbf{F} \\
        \midrule
        \textbf{Throughput} & 18.35 & 6.64 & 6.78 & 17.19 & 8.16 & 11.73 \\
        \bottomrule
    \end{tabular}%
    }
\end{table}

\subsubsection{Exploring the Impact of Device Network Bandwidth}

\setlength{\textfloatsep}{0pt}
\begin{figure}[htbp]
\centerline{\includegraphics[width=\linewidth, height=3.5cm]{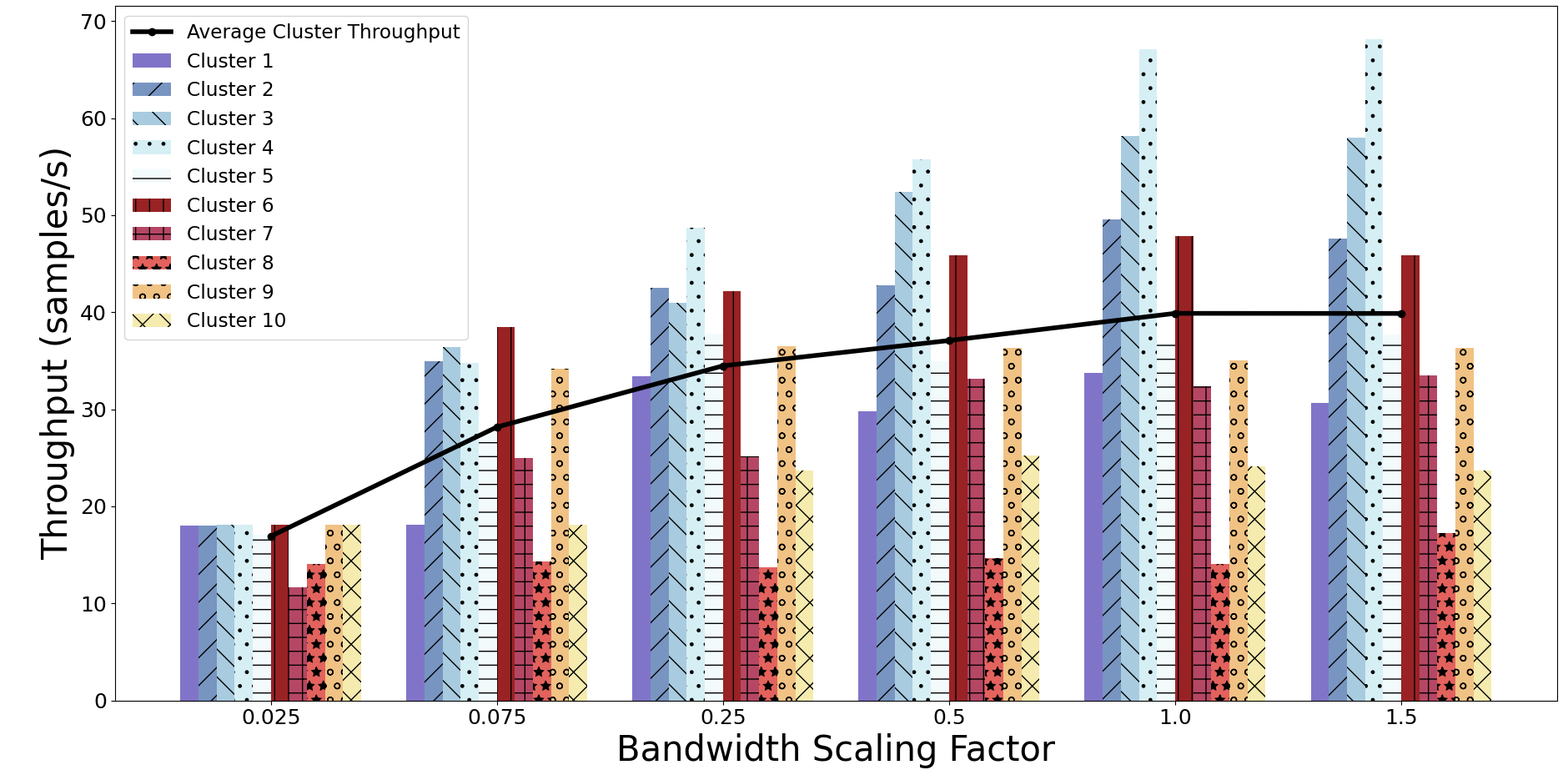}}
\caption{Cluster throughput with bandwidth scaling factors.}
\label{bandwidth_scaling}
\Description{bandwidth_scaling}
\end{figure}

We explored the impact of device bandwidth on pipeline throughput by scaling, where bandwidth settings presented in Table \ref{tab:device_config} took the scaling factor of 1.0 as the standard. Fig.~\ref{bandwidth_scaling} shows that insufficient bandwidth (e.g., scaling factor 0.025) constrained intermediate data transfer efficiency, creating a system bottleneck. We found that in such cases, GA-DPHDS tended to allocate layers to the device with the highest computing power to minimize transmission costs, reducing pipeline parallelism. For instance, in Clusters 1–4, all layers were assigned to device A, resulting in low pipeline throughput. As the scaling factor increased, the average throughput improved (black line), with clusters containing more homogeneous devices showing higher sensitivity to bandwidth, while clusters 8–10 were less affected. Once bandwidth became sufficient (e.g., scaling factor 1.5), it ceased to be a bottleneck, and device computational capacity (cpu) emerged as the new limiting factor. These results demonstrate the adaptability of GA-DPHDS under varying bandwidth.

\subsubsection{Optimization Ablation}

\setlength{\textfloatsep}{0pt}
\begin{figure}[htbp]
\centerline{\includegraphics[width=\linewidth,height=3.5cm]{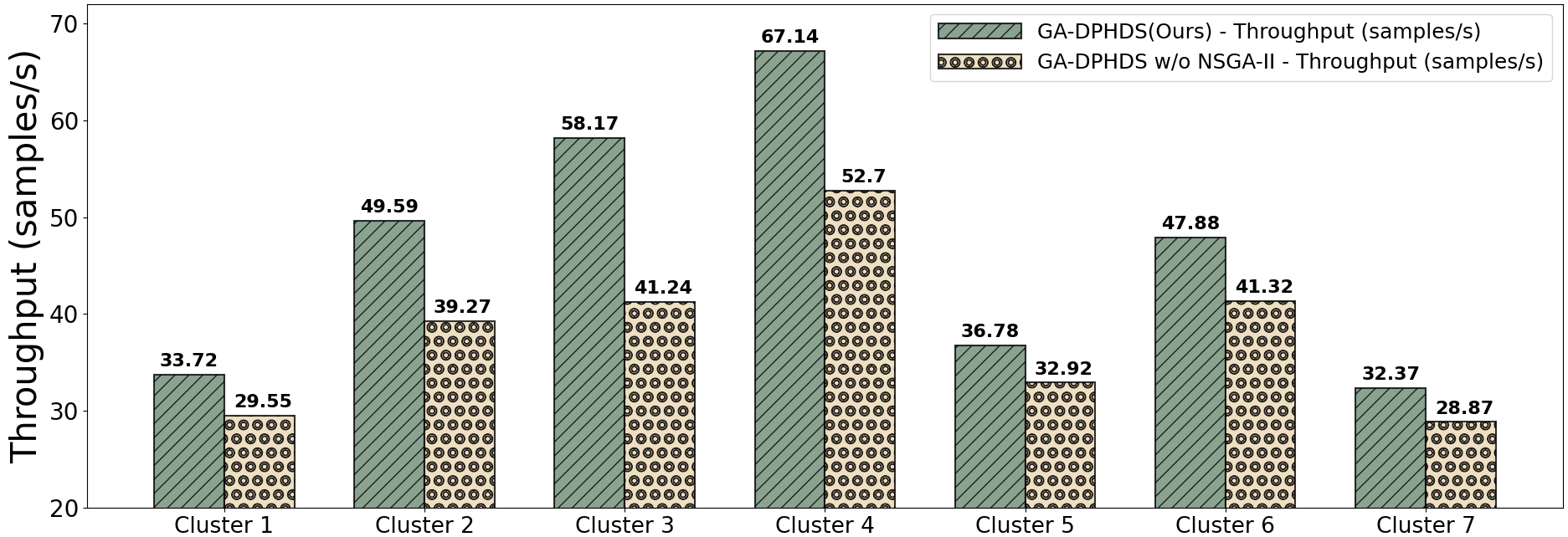}}
\caption{Ablation study: layer execution order optimization.}
\label{layer execution order optimization}
\Description{layer execution order optimization}
\end{figure}

We also examined the impact of layer execution order \( O \) on throughput, optimized via the NSGA-II genetic algorithm. Without this optimization, layer sequences were randomly arranged, adhering only to input-output dependencies. As shown in Fig.~\ref{layer execution order optimization}, optimized layer execution significantly improved throughput. In Cluster 3, GA-DPHDS achieved 58.17 samples/s, a 41.04\% increase compared to 41.24 samples/s without NSGA-II, marking the highest observed improvement. Similarly, in Cluster 4, with all the fog devices, throughput was improved by 27.4\%, demonstrating that optimized execution order enhances pipeline parallelism and system efficiency. These results highlight the criticality of execution order optimization in maximizing pipeline throughput.

\section{Conclusion and Future Work} \label{sec:conclusion}
In this work, we propose MultiGran-STGCNFog, an efficient GNN inference system with a novel traffic forecasting model, which extracts spatiotemporal features across various spatial and temporal scales, and the distributed pipeline-parallel architecture of it enables high-performance inference throughput leveraging heterogeneous fog devices. Specifically, the dynamic mechanism and multi-granular feature fusion strengthen its capability to capture long-term traffic dependencies. The proposed scheduling algorithm GA-DPHDS, brings significant throughput improvement. In the future, we might develop more delicate scheduling features such as work-stealing to further balance pipeline workload during runtime.

\bibliographystyle{ACM-Reference-Format}
\bibliography{sample-base}

\end{document}